\documentclass[final]{cvpr}
\usepackage[pagebackref=true,breaklinks=true,colorlinks,bookmarks=false]{hyperref}

\def\cvprPaperID{3270} % *** Enter the CVPR Paper ID hcere
\def\confYear{CVPR 2021}
\usepackage{times}
\usepackage{epsfig}
\usepackage{graphicx}
\usepackage{amsmath}
\usepackage{amssymb}
\usepackage{multirow}
\usepackage{booktabs}
\usepackage{bm}
\usepackage{xcolor}
\usepackage{pifont}
\usepackage{scalerel}
\usepackage{hyperref} 
\usepackage{enumitem}
\usepackage[toc,page]{appendix}

\definecolor{brightgreen}{rgb}{0.4, 1.0, 0.0}
\definecolor{chartreuse}{rgb}{0.5, 1.0, 0.0}
\definecolor{gain}{rgb}{0.224, 0.710, 0.290}
\newcommand{\colorpm}[1]{_{\footnotesize{\pm#1}}}
\newcommand{\gain}[1]{\textbf{\color{gain}{(+#1)}}}
\newcommand{\xmark}{\ding{55}}

\usepackage{subcaption}
\begin{document}

%%%%%%%%% TITLE
\title{DER: Dynamically Expandable Representation for Class Incremental Learning}

%\author{First Author\\
%Institution1\\
%Institution1 address\\
%{\tt\small firstauthor@i1.org}
%% For a paper whose authors are all at the same institution,
%% omit the following lines up until the closing ``}''.
%% Additional authors and addresses can be added with ``\and'',
%% just like the second author.
%% To save space, use either the email address or home page, not both
%\and
%Second Author\\
%Institution2\\
%First line of institution2 address\\
%{\tt\small secondauthor@i2.org}
%}
\author{Shipeng Yan\textsuperscript{\rm 1,3,4}\thanks{Both authors contributed equally. This work was supported by Shanghai NSF Grant (No. 18ZR1425100)} 
	\quad Jiangwei Xie\textsuperscript{\rm 1}$^{*}$
	\quad Xuming He\textsuperscript{\rm 1,2}\\
	\textsuperscript{\rm 1}School of Information Science and Technology, ShanghaiTech University \quad \\
	\textsuperscript{\rm 2}Shanghai Engineering Research Center of Intelligent Vision and Imaging\\
	\textsuperscript{\rm 3}Shanghai Institute of Microsystem and Information Technology,
	Chinese Academy of Sciences\\
	\textsuperscript{\rm 4}University of Chinese Academy of Sciences\\
	\{yanshp, xiejw, hexm\}@shanghaitech.edu.cn
}

\maketitle
\pagestyle{empty}
\thispagestyle{empty}

%%%%%%%%% ABSTRACT
\begin{abstract}
    We address the problem of class incremental learning, which is a core step towards achieving adaptive vision intelligence.
    %Despite recent successes of deep neural networks, it remains challenging to achieve effective class incremental learning in visual recognition.
    In particular, we consider the task setting of incremental learning with limited memory and aim to achieve better stability-plasticity trade-off. %Existing work typically are either restrictive for learning new classes or susceptible to forgetting old knowledge. 
    To this end, we propose a novel two-stage learning approach that utilizes a dynamically expandable representation for more effective incremental concept modeling. Specifically, at each incremental step, we 
    freeze the previously learned representation and augment it with additional feature dimensions from a new learnable feature extractor. This enables us to integrate new visual concepts with retaining learned knowledge.
    We dynamically expand the representation according to the complexity of novel concepts by introducing a channel-level mask-based pruning strategy. 
    Moreover, we introduce an auxiliary loss to encourage the model to learn diverse and discriminate features for novel concepts.
    We conduct extensive experiments on the three class incremental learning benchmarks and our method consistently outperforms other methods with a large margin.\footnote{Code is available at \url{https://github.com/Rhyssiyan/DER-ClassIL.pytorch.}}
\end{abstract}
% freezes previously learned representation and expands the representation via creating novel feature extractor to enable the flexibility of model to learn novel features.
% dynamically learn the compact representation via introducing 

\vspace{-1mm}
\section{Introduction}%~\label{sec:intro}

%% Background Introduction(Setting, Application)
%% Class Incremental Learning is an important problem and has lots of applications.
% Human
Human can easily accumulate visual knowledge from past experiences and incrementally learn novel concepts.
%The goal of class incremental learning in visual cognition is to mimic human intelligence by designing an memory-efficient algorithm that is able to learn novel concepts in a sequential manner and eventually to perform well on all observed classes.
Inspired by this, the problem of class incremental learning aims to design algorithms that can learn novel concepts in a sequential manner and eventually perform well on all observed classes.
% mod: not mentioned specifically in previous paper
% Applications
Such capability is indispensable for many real-world applications such as the intelligent robot~\cite{thrun1995lifelong}, face recognition~\cite{li2017incremental} and autonomous driving~\cite{pierre2018incremental}.
However, achieving human-level incremental learning remains challenging for modern visual recognition systems.

% Currently, Neural nets perform bad to incrementally learn novel concepts.

%Despite recent success of deep learning, 
\begin{figure}[t]
	% \begin{minipage}{0.25\textwidth}
	% \end{minipage}
	% \begin{minipage}{0.25\textwidth}
	\centering
    \includegraphics[width=0.4\textwidth,height=0.3\textwidth]{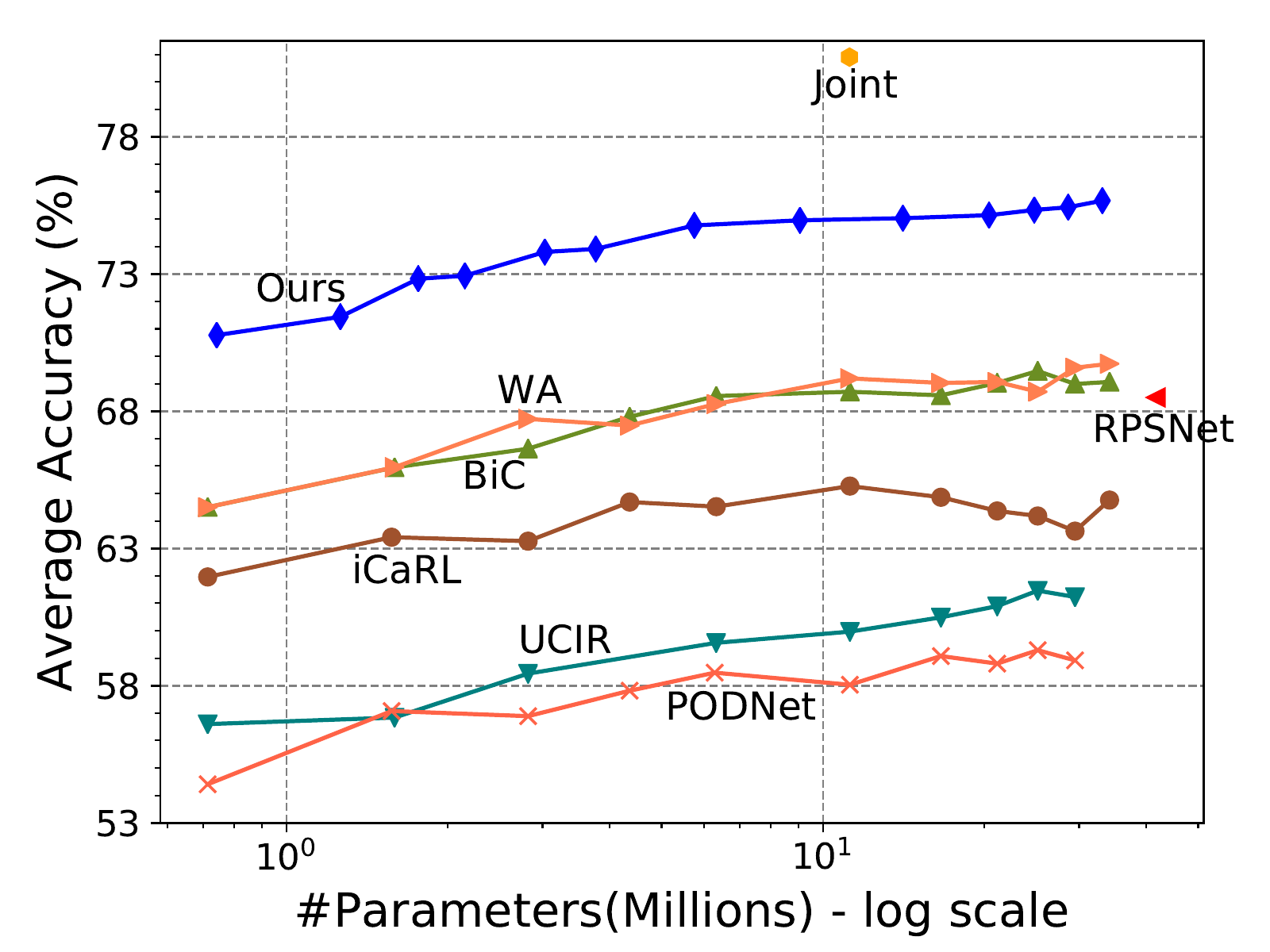}
	% \end{minipage}
	% \end{minipage}
	\vspace{-2mm}
	\caption{\fontsize{10}{12}\selectfont{The average incremental accuracy for different model size. We compare our model with prior methods (WA\cite{zhao2020wa}, BiC\cite{hou2019learning}, RPSNet\cite{rajasegaran2019random}, iCaRL\cite{rebuffi2017icarl}, UCIR\cite{hou2019learning}, PODNet\cite{douillard2020podnet}) and the model trained on all the data (Joint) on the experiment CIFAR100-B0 of 10 steps.}}
	\label{fig:cifar_model_size}
\end{figure}

% \textbf{Left} shows the comparison of classifier learning methods in which the determination of decision boundary of these methods is based on the same state-of-the-art representation.
% \emph{Baseline1 (BL1)} finetunes the classifier with available data. 
% \emph{C.U.B} is the classifier upperbound. 
% \textbf{Right} compares the quality of representation for various methods where the classifier is finetuned on all observed data based on their corresponding representation.
% \emph{Baseline2 (BL2)} means the representation is finetuned on available data with only cross entropy loss.
% \emph{Joint} means the representation is learned on all seen data.}}

% including \emph{Baseline2}, \emph{Distillation}, \emph{Ours} and \emph{Joint} 
% \emph{Baseline1}, \emph{BiC}, \emph{WA} and \emph{Classifier Upperbound(C.U.B)} 
%% Challenges(From Representation and Decision Boundary Perspective)
% mod: Intro of memory

There has been much effort attempting to address the incremental learning in literature~\cite{yu2020semantic,ostapenko2019learning,rebuffi2017icarl,castro2018eeil,hou2019learning,wu2019bic,zhao2020wa}. Among them, perhaps the most effective strategy is to keep a memory buffer that stores part of observed data for the rehearsal~\cite{Robins93a,Robins95} in future. 
However, due to the limited size of data memory, such the incremental learning method still faces several typical challenges in the general continual learning task. In particular, it requires a model to effectively incorporate novel concepts without forgetting the existing knowledge, which is also known as \textit{stability-plasticity dilemma}~\cite{grossberg2013adaptive}.
In detail, excessive plasticity often causes large performance degradation of the old categories, referred to as catastrophic forgetting~\cite{FrenchC02}. On the contrary, excessive stability impedes the adaptation of novel concepts.

Most existing works attempt to achieve a trade-off between stability and plasticity by gradually updating data representation and class decision boundary for increasingly larger label spaces. 
For instance, regularization methods~\cite{chaudhry2018riemannian} penalize the change of important weights of previously learned models, while knowledge distillation~\cite{rebuffi2017icarl,castro2018eeil,hou2019learning,douillard2020podnet,tao2020tpcil} preserves the network output with available data, and structure-based methods~\cite{rajasegaran2019random,abatiTBCCB2020ccgn} keep old parameters fixed when allocating more for new categories. Nevertheless, all those methods either sacrifice the model plasticity for the stability, or are susceptible to forgetting due to feature degradation of old concepts. As shown in Figure~\ref{fig:cifar_model_size}, a large performance gap still exists between the model (\textit{Joint}) trained on all data and previous state-of-the-art models.  

In this work, we aim to address the above weaknesses and achieve a better stability-plasticity trade-off in the class incremental learning. To this end, we adopt a two-stage learning strategy, decoupling the adaptation of feature representation and final classifier head \textit{(or classifier for short)} of a deep network~\cite{KangXRYGFK20decouple}. Within this framework, we propose a novel data representation, referred to as \textit{super-feature}, capable of increasing its dimensionality to accommodate new classes. Our main idea is to freeze the previously learned representation and augment it with additional feature dimensions from a new learnable extractor in each incremental step. This enables us to retain the existing knowledge and provides enough flexibility to learn novel concepts.
Moreover, our super-feature is expanded dynamically based on the complexity of novel concepts to maintain a compact representation.

% Model Architecture
To achieve this, we develop a modular deep classification network composed of a super-feature extractor network and a linear classifier. 
Our super-feature extractor network consists of multiple feature extractors with varying sizes, one for each incremental step. Specifically, at a new step, we expand the super-feature extractor network with a new feature extractor while keeping the parameters of previous extractors frozen. The features generated by all the extractors are concatenated together and fed into the classifier for the class prediction. 

We train the new feature extractor and the classifier on the memory and the newly incoming data. To encourage the new extractor to learn diverse and discriminative features for new classes, we design an auxiliary loss on distinguishing new and old classes. 
Additionally, to remove the model redundancy and learn the compact features for novel classes, we apply a differentiable channel-level mask-based pruning method that dynamically prunes the network according to the difficulty of novel concepts.
%In detail, we introduce a sparsity loss, which promotes the model to use fewer filters.
Finally, given the updated representation, we freeze the super-feature extractor and finetune the classifier on a balanced training subset to solve the class imbalance problem~\cite{wu2019bic,zhao2020wa}.

%In order to address aforementioned weaknesses, we propose a simple but effective dynamic structure solution that concatenation of freezed features for old concepts and compact features for novel concepts on growed novel network.
%By combining the recent works in determination of decision boundary, we find that the discrimination ability within old classes degrades slowly with the growth of feature dimension.
%% Achieve transfer
%We concat previous features and initialize the novel network to achieve backward and forward transfer.
% Previous Dynamic architecture method:
% rpsnet: additive features Dynamic Architectural -> Keep Representation

We validate our approach on three commonly used benchmarks, including CIFAR-100, ImageNet-100, and ImageNet-1000 datasets.
The empirical results and the ablation study demonstrate the superiority of our method over prior state-of-the-art approaches. 
Interestingly, we also find that our method could achieve positive backward and forward transfer between steps.
The main contributions of our work are three-fold:
\begin{itemize}[noitemsep,topsep=0pt]
\item To achieve better stability-plasticity trade-off, we develop a dynamically expandable representation and a two-stage strategy for the class incremental learning. 
\item We propose an auxiliary loss to promote the newly added feature module to learn novel classes effectively and a model pruning step to learn compact features.
\item Our approach achieves the new state of the art performance on all three benchmarks under a wide range of model complexity, as shown in Figure~\ref{fig:cifar_model_size}. 
\end{itemize}

\section{Related Work}\label{sec:realted}
Class incremental learning aims to learn new classes continuously.
Some works~\cite{yu2020semantic,ostapenko2019learning} try to solve the problem with no access to previously seen data.
However, prevalent approaches are based on the rehearsal strategy with limited data memory, which can be mainly analyzed from representation learning and classifier learning.

%Furthermore, the methods based on rehearsal strategy can be mainly analyzed from representation learning and classifier learning.
%Class incremental learning has attracted much attention recently, where most proposed methods can be mainly analyzed from representation learning and classifier learning.
\vspace{-5mm}
\paragraph{Representation Learning}
Current works can be mainly divided into the following three categories.
% Regularization method
\emph{Regularization-based} methods~\cite{kirkpatrick2017overcoming, zenke2017continual, lee2017overcoming, chaudhry2018riemannian, aljundi2018memory} adopt Maximum a Posterior estimation to expect small changes in the important parameters and update the posterior of model parameters sequentially.
%nguyen2018variational
% EWC\cite{kirkpatrick2017overcoming} estimates the weight importance through Fisher information matrix.
% EWC++\cite{chaudhry2018riemannian}
% SI\cite{zenke2017continual} uses the path integral over the optimization trajectory.
% MAS\cite{aljundi2018memory} utilizes the gradients of the L2-normalized network output.
However, its intractable computation typically requires approximations with a strong model assumption. For example, EWC~\cite{kirkpatrick2017overcoming} uses Laplace approximation, which assumes weights falling into a local region of the optimal weights of last step. This severely restricts the model capacity to adapt to novel concepts.
%These methods tend to perform poorly in class incremental learning.

% Distillation methods
\emph{Distillation-based} methods~\cite{rebuffi2017icarl,zhao2020wa,wu2019bic,hou2019learning,castro2018eeil,douillard2020podnet, tao2020tpcil} use knowledge distillation~\cite{hinton15distill} to maintain the representation.
iCaRL~\cite{rebuffi2017icarl} and EE2L~\cite{castro2018eeil} compute the distillation loss on the network outputs.
UCIR~\cite{hou2019learning} uses normalized feature vectors to apply the distillation loss instead of the prediction of the network.
PODNet~\cite{douillard2020podnet} uses a spatial-based distillation loss to restrict the change of model. 
TPCIL~\cite{tao2020tpcil} makes the model preserve the topology of CNN's feature space.
The performance of knowledge distillation depends on the quality and quantity of saved data. 

% \modify{is the most-widely used technique in class incremental learning, which retains the knowledge of learned concepts by making the network outputs unchanged. iCaRL\cite{rebuffi2017icarl} and EE2L\cite{castro2018eeil} use knowledge distillation to keep the output of classifier the same as old model. UCIR\cite{hou2019learning} maximizes the cosine distance between the normalized features instead of the prediction extracted by old model and current model.}

% Dynamic Network structure
\emph{Structure-based} methods~\cite{mallya2018piggyback,hung2019cpg,serra2018overcoming,li2019learn,mallya2018packnet,fernando2017pathnet,yoon2018lifelong,rajasegaran2019random,abatiTBCCB2020ccgn,li2019learn} keep the learned parameters related to previous classes fixed and allocate new parameters in different forms such as unused parameters, additional networks to learn novel knowledge.
CPG~\cite{hung2019cpg} proposes a compaction and selection/expansion mechanism that prunes the deep model and expands the architecture alternatively with selectively weight sharing.
However, most structure-based~\cite{mallya2018piggyback,hung2019cpg,serra2018overcoming,li2019learn,mallya2018packnet,fernando2017pathnet,yoon2018lifelong} methods are designed for task continual learning, which needs task identity during inference.
For class incremental learning, RPSNet~\cite{rajasegaran2019random} proposes a random path selection algorithm that progressively chooses optimal paths as sub-network for the new classes. %while encouraging parameter sharing between paths selected at different steps.
CCGN~\cite{abatiTBCCB2020ccgn} equips each convolutional layer with task-specific gating modules to select filters to apply on the given input and uses a task predictor to choose the gating modules in inference.

\vspace{-5mm}
\paragraph{Classifier Learning}
Class-imbalance problem is the main challenge for classifier learning due to limited memory.
%The methods can be divided into one-stage training and two-stage training which separates the classifier training as a individual stage.
Some works like LWF.MC\cite{rebuffi2017icarl}, RWalk\cite{chaudhry2018riemannian}  train the extractor and classifier jointly within one-stage training. %feature and RPSNet\cite{rajasegaran2019random}
By contrast, recently, there are many works to solve the class imbalance problem by introducing an independent classifier learning stage after representation learning.
EEIL\cite{castro2018eeil} finetunes the classifier on a balanced training subset.
BiC\cite{wu2019bic} adds a bias correction layer to correct the model's outputs, where the layer is trained on a separate validation set.
WA\cite{zhao2020wa} corrects the biased weights by aligning the norms of the weight vectors for new classes to those for old classes.

\vspace{-5mm}
\paragraph{Discussion}
Our work is a structure-based method and the most similar work to ours are RPSNet and CCGN.
%RPSNet tends to gradually forget the learned concepts by summing the features from paths of different steps at each stage.
RPSNet cannot retain the intrinsic structure of each old concept and tends to gradually forget the learned concepts by summing the previously learned features and the newly learned features at each ConvNet stage.
In CCGN, the learned representation may slowly degrade over steps as only the parameters of part of layers are frozen.
By contrast, we keep the previously learned representation fixed and augment it with novel features parameterized by a new feature extractor.
%By contrast, we keep the previously learned representation fixed and learn the novel features by adapting the new feature subnet.
This enables us to preserve the intrinsic structure of old concepts in the subspace of previously learned representation, and re-use the structure via the final classifier to mitigate forgetting. %, which can be obtained from the recent advances on classifier learning.

\vspace{-0.2mm}
\section{Methods}\label{sec:methods}
In this section, we present our approach to the problem of class incremental learning, aiming to achieve a better trade-off between stability and plasticity. To this end, we propose a dynamically expandable representation (DER) that incrementally augments previously learned representation with novel features and a two-stage learning strategy. 
%by creating an additional feature extractor. We adaptively expand the representation according to the complexity of new categories, and re-learn the classifier head given the new representation. 

%We also design an auxiliary loss to promote learning diverse and discriminative features for novel classes.
%To improve the model efficiency, we dynamically expand the representation according to the complexity of new categories by introducing the channel-level mask-based pruning method.

Below we first present the formulation of class incremental learning and an overview of our method in Sec.~\ref{subsec:formulation}.
Then we introduce the expandable representation learning and its loss function in Sec.~\ref{subsec:training}. 
%we describe the dynamic expansion of representation via pruning in Sec.~\ref{subsec:compression}.
After this, we describe the dynamic expansion of our representation in Sec.~\ref{subsec:compression} and the second stage of classifier learning in Sec.~\ref{subsec:cls}.

\begin{figure*}[t]
	\centering
	\includegraphics[width=.85\textwidth]{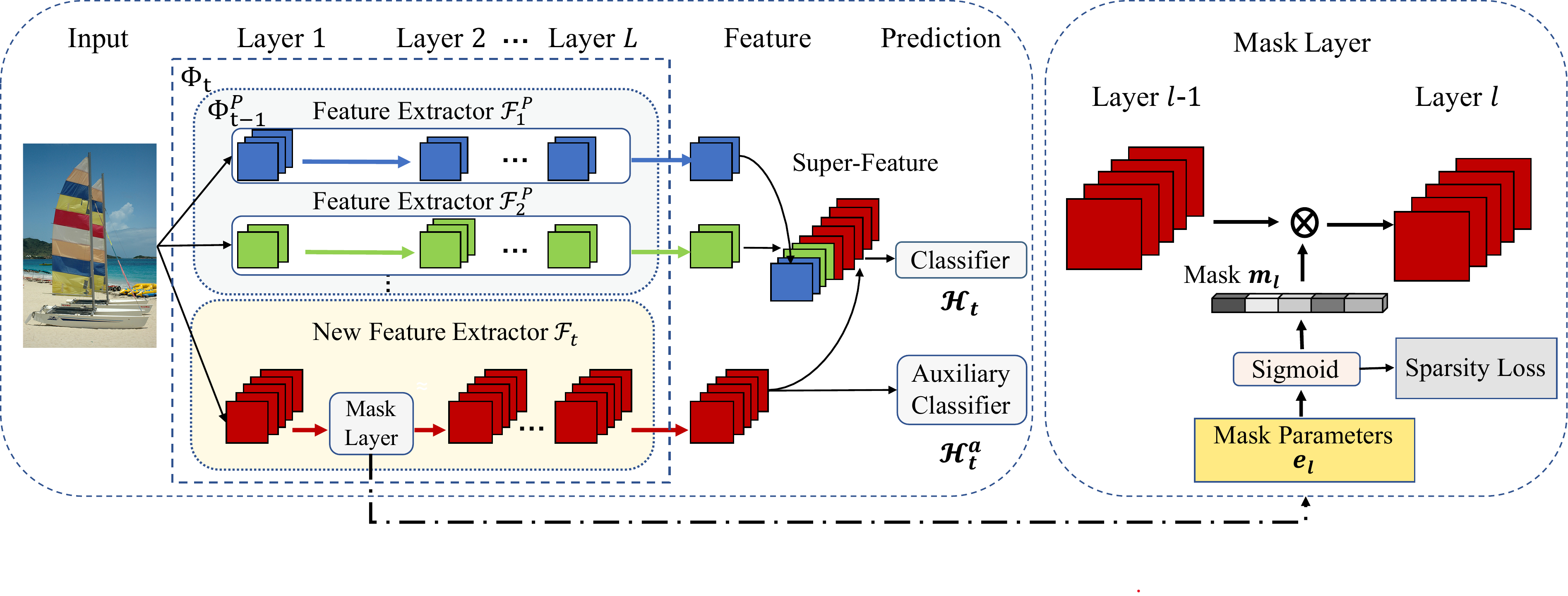}
	\vspace{-7mm}
	\caption{Dynamically Expandable Representation Learning. At step $t$, the model is composed of super-feature extractor $\Phi_t$ and classifier $\mathcal{H}_t$, where $\Phi_t$ is built by expanding the existing super-feature extractor $\Phi_{t-1}^P$ with new feature extractor $\mathcal{F}_t$.
We also use an auxiliary classifier to regularize the model. Besides, the layer-wise channel-level mask is jointed learned with the representation, which is used to prune the network after the learning of model.}
	\label{fig:model_arch}
%	\vspace{-2mm}
	\end{figure*} 

\subsection{Problem Setup and Method Overview}~\label{subsec:formulation}
% In this work, we aim to address the class incremental learning problem for the image  classification task.
% The problem could be formulated as follows.
Firstly, we introduce the problem setup of class incremental learning. In contrast to task incremental learning, class incremental learning does not require task id during inference.
Specifically, during the class incremental learning, the model observes a stream of class groups $\{\mathcal{Y}_t\}$ and their corresponding training data $\{\mathcal{D}_t\}$.
Particularly, the incoming dataset $\mathcal{D}_t$ at step $t$ has a form of $(\bm{x^t_i}, y^t_i)$ where $\bm{x^t_i}$ is the input image and $y^t_i\in \mathcal{Y}_t$ is the label within the label set $\mathcal{Y}_t$. 
The label space of the model is all seen categories $\tilde{\mathcal{Y}}_t= \cup_{i=1}^t \mathcal{Y}_i$ and the model is expected to predict well on all classes in $\tilde{\mathcal{Y}}_t$.

%In the class incremental learning scenario, the data arrives in an incremental fashion and hence it is infeasible to optimize all observed classes jointly.
Our method adopts the rehearsal strategy, which saves a part of data as the memory $\mathcal{M}_t$ for future training.
For the learning of step $t$, we decouple the learning process into two sequential stages as follows.

\vspace{1mm}
\noindent\textit{1) Representation Learning Stage.}
To achieve better trade-off between stability and plasticity, we fix the previous feature representation and expand it with a new feature extractor trained on the incoming and memory data.
We design an auxiliary loss on the novel extractor to promote it to learn diverse and discriminative features. %feature
To improve the model efficiency, we dynamically expand the representation according to the complexity of new classes via introducing a channel-level mask-based pruning method.
% according to the complexity of new categories
%reduce the size of the new feature extractor.
The overview of our proposed representation is shown in Figure~\ref{fig:model_arch}.

\vspace{1mm}
\noindent\textit{2) Classifier Learning Stage.}
After the learning of representation, we retrain the classifier with currently available data $\tilde{\mathcal{D}}_t = \mathcal{D}_t \cup \mathcal{M}_t$ at step $t$ to deal with the class imbalance problem via adopting the balanced finetuning method in \cite{castro2018eeil}.
% proposed in 
%we adopt the balanced finetuning method proposed in \cite{castro2018eeil} to deal with the class imbalance problem \more{with currently available data $\tilde{\mathcal{D}}_t = \mathcal{D}_t \cup \mathcal{M}_t$ at step $t$.}
%Specifically, we train the classifier with a class-balanced subset sampled from currently available data $\tilde{\mathcal{D}}_t = \mathcal{D}_t \cup \mathcal{M}_t$ at step $t$.

\subsection{Expandable Representation Learning}~\label{subsec:training}
We first introduce our expandable representation.
At step $t$, our model is composed of a super-feature extractor $\Phi_t$ and the classifier $\mathcal{H}_t$.
The super-feature extractor $\Phi_t$ is built by expanding the feature extractor $\Phi_{t-1}$ with a newly created feature extractor $\mathcal{F}_t$.
Specifically, given an image $\bm{x} \in \tilde{\mathcal{D}}_{t}$, the feature $\bm{u}$ extracted by $\Phi_t$ is obtained by concatenation as follows
\begin{align}\label{eq:1}
	\bm{u} = \Phi_t(\bm{x}) = [\Phi_{t-1}(\bm{x}), \mathcal{F}_t(\bm{x})]
\end{align}
Here we reuse the previous $\mathcal{F}_{1},\dots,\mathcal{F}_{t-1}$ and encourage the new extractor $\mathcal{F}_t$ to learn only novel aspect of new classes.
The feature $\bm{u}$ is then fed into the classifier $\mathcal{H}_t$ to make prediction as follows
\begin{align}\label{eq:2}
	p_{\mathcal{H}_{t}}(\bm{y}|\bm{x}) = \text{Softmax}(\mathcal{H}_t(\bm{u}))
\end{align}
Then the prediction $\hat{y} = \arg \max p_{\mathcal{H}_{t}}(\bm{y}|\bm{x})$, $ \hat{y} \in \tilde{\mathcal{Y}_t}$.
The classifier is designed to match its new input and output dimensions for step $t$.
The parameters of $\mathcal{H}_t$ for the old features are inherited from $\mathcal{H}_{t-1}$ to retain old knowledge and its newly added parameters are randomly initialized.
%better memorize the old classes
%$\mathcal{H}_t$ $\mathcal{H}_{t-1}$  compared to 

To reduce catastrophic forgetting, we freeze the learned function $\Phi_{t-1}$ at step $t$, as it captures the intrinsic structure of previous data. 
In detail, the parameters of last step super-feature extractor $\theta_{\Phi_{t-1}}$ and the statistics of Batch Normalization\cite{IoffeS15bn} are not updated.
Besides, we instantiate $\mathcal{F}_t$ with $\mathcal{F}_{t-1}$ as initialization to reuse previous knowledge for fast adaptation and forward transfer.
% It is worth noting that we instantiate $\mathcal{F}_t$ with $\mathcal{F}_{t-1}$ as initialization.

% At step $t$, we first freeze the hyper-feature extractors inherited from last last to overcome catastrophic forgetting, as it captures the intrinsic structure of previous data. Then we built $\Phi_{t}$ by extending $\Phi_{t-1}$ with a newly created feature extractor $\mathcal{F}_t$, which provides the flexiblility for model to recognize novel concepts.
% In detail, the parameters of last step hyper-feature extractor $\theta_{\Phi_{t-1}}$ and the statistics of Batch Normalization\cite{IoffeS15bn} are not updated and $\mathcal{F}_t$ is initialized from $\mathcal{F}_{t-1}$.
% After builting $\Phi_{t}$, given an image $\bm{x} \in \tilde{\mathcal{D}}_{t}$, the hyper-feature $\bm{u}$ extracted by $\Phi_t$ is obtained by concatenation as follows
% \begin{align}
% 	\bm{u} = \Phi_t(\bm{x}) = [\Phi_{t-1}(\bm{x}), \mathcal{F}_t(\bm{x})]
% \end{align}
% The hyper-feature $\bm{u}$ is then fed into the classifier head $\mathcal{H}_t$ to make prediction as follows
% \begin{align}
% 	p_{\mathcal{H}_{t}}(\bm{y}|\bm{x}) = \text{Softmax}(\mathcal{H}_t(\bm{u}))
% \end{align}
% where the predicted label $\hat{\bm{y}} = \arg \max p_{\mathcal{H}_{t}}(\bm{y}|\bm{x}), \hat{\bm{y}} \in \tilde{\mathcal{Y}_t}$

% View it from Bayesian Perspective
We can shed the light on the problem from the perspective of estimating the prior distribution $p(\theta_{\Phi_t}|\mathcal{D}_{1:t-1})$ given the previous data $\mathcal{D}_{1:t-1}$.
Unlike previous regularization methods like EWC, we do not assume the prior distribution for $t$-th step is unimodal, which restricts the model flexibility and is typically not the case in practice.
For our method, the model expands with new parameters by creating a separate feature extractor $\mathcal{F}_t$ for incoming data and take a uniform distribution as the prior distribution $p(\theta_{\mathcal{F}_t}|\mathcal{D}_{1:t-1})$ which provides enough flexibility for the model to adapt to novel concepts.
Meanwhile, for simplicity, we approximate the prior distribution $p(\theta_{\Phi_{t-1}}|\mathcal{D}_{1:t-1})$ on the old parameters $\theta_{\Phi_{t-1}}$ as the Dirac distribution, which maintains the information learned on $\mathcal{D}_{1:t-1}$.
By integrating two prior distribution assumptions on $p(\theta_{\Phi_{t-1}}|\mathcal{D}_{1:t-1})$ and $p(\theta_{\mathcal{F}_t}|\mathcal{D}_{1:t-1})$, we have more flexibility in achieving a better stability-plasticity trade-off.
% \paragraph{Representation Learning}

% \paragraph{Model Architecture/Forward}
% $a^l_t = \sigma(se_t)$
\vspace{-5mm}
\paragraph{Training Loss}
% Goal
We learn the model with cross-entropy loss on memory and incoming data as follows
\vspace{-1mm}
\begin{align}\label{eq:3}
	\mathcal{L}_{\mathcal{H}_t} = - \frac{1}{|\tilde{\mathcal{D}}_t|} \sum_{i=1}^{|\tilde{\mathcal{D}}_t|} \log(p_{\mathcal{H}_t}(y=y_i|\bm{x_i})))
\vspace{-1mm}
\end{align}
where $\bm{x_i}$ is image and $y_i$ is the corresponding label.

To enforce the network to learn the diverse and discriminative features for novel concepts, we further develop an auxiliary loss operating on the novel feature $\mathcal{F}_t(\bm{x})$.
Specifically, we introduce an auxiliary classifier $\mathcal{H}_t^a$, which predicts the probability $p_{\mathcal{H}_t^a}(\bm{y}|\bm{x})=\text{Softmax}(\mathcal{H}_t^a(\mathcal{F}_t(\bm{x}))$.
To encourage the network to learn features to discriminate between old and new concepts, the label space of $\mathcal{H}^a_t$ is $|\mathcal{Y}_t|+1$ including the new category set $\mathcal{Y}_t$ and the other class by treating all old concepts as one category.
Thusly, we introduce the auxiliary loss and obtain the expandable representation loss as follows
\begin{align}
\mathcal{L}_{\text{ER}} = \mathcal{L}_{\mathcal{H}_t} + \lambda_a\mathcal{L}_{\mathcal{H}^a_t}
\end{align}
where $\lambda_a$ is the hyper-parameter to control the effect of the auxiliary classifier.
It is worth noting that $\lambda_a$$=$$0$ for first step $t=1$.

\subsection{Dynamical Expansion}\label{subsec:compression}
% Goal of compression
To remove the model redundancy and maintain a compact representation, we dynamically expand the super-feature according to the complexity of novel concepts.
Specifically, we adopt a differentiable channel-level mask-based method to prune filters of the extractor $\mathcal{F}_t$, in which the masks are learned with the representation jointly. 
After the learning of the mask, we binarize the mask and prune the feature extractor $\mathcal{F}_t$ to obtain the pruned network $\mathcal{F}^P_t$. 

\vspace{-3mm}
\paragraph{Channel-level Masks} 
% $M$ means the mask is applied on the feature extractor and there is no sharable parameters between $\Phi_{t-1}$ and $\mathcal{F}^M_t$.
Our pruning method is based on differentiable channel-level masks, which is adapted from HAT \cite{serra2018overcoming}.
For the novel feature extractor $\mathcal{F}_t$, the input feature map of  convolutional layer $l$ for a given image $\bm{x}$ is denoted as $\bm{f_l}$.
We introduce the channel mask $\bm{m_{l}} \in \mathbb{R}^{c_{l}}$ to control the size of layer $l$ where  $m^i_l\in [0,1]$ and $c_l$ is the number of channels of layer $l$. 
$\bm{f_l}$ is modulated with the mask as follows
\begin{align}
	\bm{f'_{l}} = \bm{f_{l}} \odot \bm{m_{l}}
\end{align}
where $\bm{f'_{l}}$ is the masked feature map, $\odot$ means channel-level multiplication.
To make the value of $\bm{m_l}$ fall into the interval $[0,1]$, the gating function is adopted as follows
\begin{equation}
	\bm{m_l} = \sigma(s\bm{e_l})
\end{equation}
where $\bm{e_l}$ means learnable mask parameters, the gating function $\sigma(\cdot)$ uses the sigmoid function in this work and $s$ is the scaling factor to control the sharpness of the function.
With such a mask mechanism, the super-feature $\tilde{u}$ of step $t$ can be rewritten as 
\begin{align}
	\tilde{\bm{u}} = \Phi^{P}_t(\bm{x}) = [\mathcal{F}^{P}_1(\bm{x}), \mathcal{F}^{P}_2(\bm{x}), ..., \phi_t(\bm{x})]
\end{align}
During training, $\phi_t(\bm{x})$ is $\mathcal{F}_{t}(\bm{x})$ with the soft masks.
For inference, we assign $s$ a large value to binarize masks and obtain the pruned network $\mathcal{F}^{P}_{t}$, and $\phi_t(\bm{x})=\mathcal{F}^P_t(\bm{x})$ 

%, where $P$ means the feature extractor is pruned.
\vspace{-3mm}
\paragraph{Mask Learning} 
During a epoch, a linear annealing schedule is applied for $s$ as follows
\begin{equation}
	s = \frac{1}{s_{\text{max}}} + (s_{\text{max}} - \frac{1}{s_{\text{max}}}) \frac{b - 1}{B - 1}
\end{equation}
where $b$ is the batch index, $s_{\text{max}}$$\gg$$1$ is the hyper-parameter to control the schedule, $B$ is the number of batches in one epoch.
The training epoch starts with all channels activated in a uniform way.
Then the mask is progressively binarized with the increasing of batch index within a epoch.

% gradient compensation
One of the problems of the sigmoid function is that the gradient is unstable due to the $s$ schedule. We compensate the gradient $\bm{g_{e_l}}$ with respect to $\bm{e_l}$ to remove the influence of $s$ as follows

\begin{equation}
	\bm{g'_{e_l}} = \frac{\sigma(\bm{e_l}) [1 - \sigma(\bm{e_l})] }{s \sigma(s\bm{e_l}) [1 - \sigma (s \bm{e_{l}})] } \bm{g_{e_l}}
\end{equation}
where $\bm{g'_{e_l}}$ is the compensated gradient.
\vspace{-3mm}
\paragraph{Sparsity Loss}
At every step, we encourage the model to maximally reduce the number of parameters with a minimal performance drop.
Motivated by this, we add a sparsity loss based on the ratio of used weights in all available weights.
\begin{equation}
	\mathcal{L}_S =  \frac{\sum_{l=1}^{L} K_l \lVert\bm{m_{l-1}}\rVert_{1}  \lVert\bm{m_{l}}\rVert_{1}} {\sum_{l=1}^{L} K_l c_{l-1}  c_{l}}
\end{equation}
where $L$ is the number of layers, $K_l$ is the kernel size of convolution layer $l$, layer $l$$=$$0$ refers to the input image, and $\lVert \bm{m_0}\rVert_1$$=$$3$.
% Details. (1) binary mask / forward (2) special method (3) loss

After adding the sparsity loss, the final loss function is 
\begin{align}
\mathcal{L}_{\text{DER}} = \mathcal{L}_{\mathcal{H}_t} + \lambda_a \mathcal{L}_{\mathcal{H}_t^a} + \lambda_s \mathcal{L}_S
\end{align}
where $\lambda_s$ is the hyper-parameter to control the model size.

\subsection{Classifier Learning}~\label{subsec:cls}
At the representation learning stage, we re-train the classifier head in order to reduce the bias in the classifier weight introduced by the imbalanced training.
Specifically, we first re-initialize the classifier with random weights and then sample a class-balanced subset from currently available data $\tilde{\mathcal{D}}_t$. We train the classifier head only using the cross-entropy loss with a temperature $\delta$ in the Softmax~\cite{zhang2018heated}. The temperature controls the smoothness of the Softmax function to improve the margins between classes.

\section{Experiments}\label{sec:exp}
In this section, we conduct extensive experiments to validate the effectiveness of our algorithm.
Especially, we evaluate our method on CIFAR-100\cite{rebuffi2017icarl}, ImageNet-100\cite{rebuffi2017icarl} and ImageNet-1000\cite{rebuffi2017icarl} datasets with two widely used benchmark protocols.
We also perform a series of ablation studies to evaluate the importance of each component and provides more insights into our method.
Below we first start with the introduction of experiment setup and implementation details in Sec.~\ref{subsec:exp_setup}, followed by the experimental results on the CIFAR100 dataset in Sec.~\ref{subsec:cifar100}.
Then we present the evaluation results on both ImageNet-100 and ImageNet-1000 datasets in Sec.~\ref{subsec:imagenet}.
Finally, we introduce the ablation study and analysis for our method in Sec.~\ref{subsec:ablation}.

\begin{table*}[t]
\centering
%\caption{Results on CIFAR100-B0 (average over 3 runs). \emph{$\#$Paras} means the average number of parameters used during inference over steps, whose unit is million parameters. \emph{Avg} means the average accuracy ($\%$) over steps. \emph{Ours} represents our method and \emph{Ours(w/o P)} means our method without pruning.}
\fontsize{10.53}{12}\selectfont{
\resizebox{0.95\textwidth}{!}{
\begin{tabular}{l|ll|ll|ll|ll}
%\centering
\toprule[0.3mm]
\multirow{2}{*}{\textbf{Methods}} &\multicolumn{2}{c|}{\textbf{$5$ steps}}       & \multicolumn{2}{c|}{\textbf{$10$ steps}}     & \multicolumn{2}{c|}{\textbf{$20$ steps}}    & \multicolumn{2}{c}{\textbf{$50$ steps}} \\
\cmidrule{2-9}
&                                \textbf{$\#$Paras}   & \textbf{Avg}  & \textbf{$\#$Paras}  & \textbf{Avg}     & \textbf{$\#$Paras}   & \textbf{Avg} & \textbf{$\#$Paras} & \textbf{Avg}    \\
\midrule
Bound                              & $11.2$ & $80.40$               & $11.2$ & $80.41$               & $11.2$ & $81.49$               & $11.2$ & $81.74$                 \\
\midrule
iCaRL\cite{rebuffi2017icarl}       & $11.2$ & $71.14\colorpm{0.34}$ & $11.2$ & $65.27\colorpm{1.02}$ & $11.2$ & $61.20\colorpm{0.83}$ & $11.2$ & $56.08\colorpm{0.83}$ \\
UCIR\cite{hou2019learning}         & $11.2$ & $62.77\colorpm{0.82}$ & $11.2$ & $58.66\colorpm{0.71}$ & $11.2$ & $58.17\colorpm{0.30}$ & $11.2$ & $56.86\colorpm{3.74}$ \\
BiC\cite{hou2019learning}          & $11.2$ & $73.10\colorpm{0.55}$ & $11.2$ & $68.80\colorpm{1.20}$ & $11.2$ & $66.48\colorpm{0.32}$ & $11.2$ & $62.09\colorpm{0.85}$ \\
WA\cite{zhao2020wa}       & $11.2$ & $72.81\colorpm{0.28}$ & $11.2$ & $69.46\colorpm{0.29}$ & $11.2$ & $67.33\colorpm{0.15}$ & $11.2$ & $64.32\colorpm{0.28}$ \\
PODNet\cite{douillard2020podnet}   & $11.2$ & $66.70\colorpm{0.64}$ & $11.2$ & $58.03\colorpm{1.27}$ & $11.2$ & $53.97\colorpm{0.85}$ & $11.2$ & $51.19\colorpm{1.02}$ \\
RPSNet\cite{rajasegaran2019random} & $60.6$ &   $70.5$          & $56.5$  &      $68.6$               &   -   &  - & - &  - \\
\midrule
Ours(w/o P) & $33.6$ & $\textbf{76.80}\colorpm{0.79}\gain{3.7}$  & $61.6$ & $\textbf{75.36}\colorpm{0.36}\gain{5.9}$ & $117.6$ & $\textbf{74.09}\colorpm{0.33}\gain{6.76}$ & $285.6$ & $\textbf{72.41}\colorpm{0.36}\gain{8.09}$  \\
Ours        & $\textbf{2.89}$ & $\textbf{75.55}\colorpm{0.65}\gain{2.45}$ & $\textbf{4.96}$ & $\textbf{74.64}\colorpm{0.28}\gain{5.18}$ & $\textbf{7.21}$ & $\textbf{73.98}\colorpm{0.36}\gain{6.65}$ & $\textbf{10.15}$ & $\textbf{72.05}\colorpm{0.55}\gain{7.73}$ \\
\bottomrule[0.3mm]
\end{tabular}}}
\caption{\fontsize{10}{12}\selectfont{Results on CIFAR100-B0 benchmark which is averaged over three runs. \emph{$\#$Paras} means the average number of parameters used during inference over steps, which is counted by million. \emph{Avg} means the average accuracy ($\%$) over steps. \emph{Ours(w/o P)} means our method without pruning.}}
\label{tab:cifar100_base0}
%$\textbf{3}$$$$\color{gain}{(+3.7)}$
\end{table*}

\subsection{Experiment Setup and Implementation Details}\label{subsec:exp_setup}
\begin{table*}[t]
	\centering
	\fontsize{11.76}{12}\selectfont{
		\resizebox{0.83\textwidth}{!}{
			\footnotesize
			\begin{tabular}{l|ll|ll|ll}
				\toprule[0.3mm]
				\multirow{2}{*}{\textbf{Methods}}         & \multicolumn{2}{c|}{\textbf{2Steps}}   & \multicolumn{2}{c|}{\textbf{5Steps}}               & \multicolumn{2}{c}{\textbf{10Steps}}                                                                                                                          \\
				\cmidrule{2-7}
				                                 & \textbf{$\#$Paras}                     & \textbf{Avg}                                       & \textbf{$\#$Paras}                     & \textbf{Avg}                                       & \textbf{$\#$Paras}                     & \textbf{Avg}                                       \\
				\midrule
				Bound                            & $11.2$                        & $77.22$                                   & $11.2$                        & $79.89$                                   & $11.2$                        & $79.91$                                   \\
				\midrule
				iCaRL\cite{rebuffi2017icarl}     & $11.2$                        & $71.33\colorpm{0.35}$                     & $11.2$                        & $65.06\colorpm{0.53}$                     & $11.2$                        & $58.59\colorpm{0.95}$                     \\
				UCIR\cite{hou2019learning}       & $11.2$                        & $67.21\colorpm{0.35}$                     & $11.2$                        & $64.28\colorpm{0.19}$                     & $11.2$                        & $59.92\colorpm{2.4}$                      \\
				BiC\cite{hou2019learning}        & $11.2$                        & $72.47\colorpm{0.99}$                     & $11.2$                        & $66.62\colorpm{0.45}$                     & $11.2$                        & $60.25\colorpm{0.34}$                     \\
				WA\cite{zhao2020wa}     & $11.2$                        & $71.43\colorpm{0.65}$                     & $11.2$                        & $64.01\colorpm{1.62}$                     & $11.2$                        & $57.86\colorpm{0.81}$                     \\
				PODNet\cite{douillard2020podnet} & $11.2$                        & $71.30\colorpm{0.46}$                     & $11.2$                        & $67.25\colorpm{0.27}$                     & $11.2$                        & $64.04\colorpm{0.43}$                     \\
				%\modify{RPSNet\cite{rajasegaran2019random}} &      &   &     &       &     &         \\
				\midrule
				Ours(w/o P) & $22.4$          & $\textbf{74.61}\colorpm{0.52}\gain{2.14}$ & $39.2$          & $\textbf{73.21}\colorpm{0.78}\gain{5.96}$ & $67.2$          & $\textbf{72.81}\colorpm{0.88}\gain{8.77}$ \\
				Ours                             & $\textbf{3.90}$ & $\textbf{74.57}\colorpm{0.42}\gain{2.10}$ & $\textbf{6.13}$ & $\textbf{72.60}\colorpm{0.78}\gain{5.35}$ & $\textbf{8.79}$ & $\textbf{72.45}\colorpm{0.76}\gain{8.41}$ \\
				\bottomrule[0.3mm]
			\end{tabular}}}
	\caption{\fontsize{10}{12}\selectfont{Results on CIFAR100-B50 (average over 3 runs). \emph{$\#$Paras} means the average number of parameters used during inference over steps, which is counted by million. \emph{Avg} means the average accuracy ($\%$) over steps. \emph{Ours(w/o P)} means our method without pruning.}} \vspace{-3mm}
	\label{tab:cifar100_base50}
\end{table*}

\begin{table*}[t]
%	\centering
%	\hspace{mm}
\fontsize{12}{14}\selectfont{
	\begin{minipage}[h]{0.61\textwidth}
	\resizebox{1\textwidth}{!}{
			\begin{tabular}{l|ccccc|ccccc}
			\toprule[0.3mm]
			\multirow{3}{*}{\textbf{Methods}} & \multicolumn{5}{c|}{\textbf{ImageNet100-B0}} & \multicolumn{5}{c}{\textbf{ImageNet1000-B0}} \\
			\cmidrule{2-11}
			 & \multirow{2}{*}{\textbf{$\#$Paras}} & \multicolumn{2}{c}{\textbf{top-1}}  & \multicolumn{2}{c|}{\textbf{top-5}} & \multirow{2}{*}{\textbf{$\#$Paras}}   & \multicolumn{2}{c}{\textbf{top-1}} & \multicolumn{2}{c}{\textbf{top-5}} \\
			 			\cmidrule{3-6}
			\cmidrule{8-11}
			&   & \textbf{Avg}           & \textbf{Last}        & \textbf{Avg}     & \textbf{Last}   &     & \textbf{Avg}    & \textbf{Last}      & \textbf{Avg}     & \textbf{Last}      \\
			\midrule
			Bound  & $11.2$  & - & -  & - & $95.1$ & $11.2$ & $89.27$ & - & - & - \\
			\midrule
			iCaRL\cite{rebuffi2017icarl} & $11.2$ & - & - & $83.6$ & $63.8$ & $11.2$ & $38.4$ & $22.7$& $63.7$ & $44.0$ \\
			BiC\cite{hou2019learning} & $11.2$ & - & - & $90.6$ & $84.4$ & $11.2$ & - & - & $84.0$ & $73.2$\\
			WA\cite{zhao2020wa} & $11.2$ & - & - & $91.0$ & $84.1$ & $11.2$ & $65.67$ & $55.6$ & $86.6$ & $81.1$ \\
			RPSNet\cite{rajasegaran2019random} & - & - & - & $87.9$ & $74.0$ & - & - & - & - & - \\
			\midrule
			Ours(w/o P)  & $61.6$ & $\textbf{77.18}$ & $\textbf{66.70}$  & $\textbf{93.23}$ & $\textbf{87.52}$  & $61.6$ & $\textbf{68.84}$ & $\textbf{60.16}$ & $\textbf{88.17}$ & $\textbf{82.86}$  \\
			Ours & $\textbf{7.67}$ & $\textbf{76.12}$  & $\textbf{66.06}$ & $\textbf{92.79}$ & $\textbf{88.38}$  & $14.52$ & $\textbf{66.73}$ & $\textbf{58.62}$  & $\textbf{87.08}$ & $\textbf{81.89}$\\
			\bottomrule[0.3mm]
			\end{tabular}}
	\end{minipage}}
	\hspace{-1mm}
	\vspace{-0.4mm}
	\fontsize{10}{11.9}\selectfont{
	\begin{minipage}{0.4\textwidth}
	\resizebox{0.95\textwidth}{!}{
	\begin{tabular}{l|ccccc}
	\toprule[0.3mm]
	\multirow{3}{*}{\textbf{Methods}} & \multicolumn{5}{c}{\textbf{ImageNet100-B50}}  \\
			\cmidrule{2-6}
			 & \multirow{2}{*}{\textbf{$\#$Paras}} & \multicolumn{2}{c}{\textbf{top-1}}  & \multicolumn{2}{c}{\textbf{top-5}}  \\
			 \cmidrule{3-6}
			 &        & \textbf{Avg}          & \textbf{Last}    & \textbf{Avg} & \textbf{Last}   \\
			 \midrule
			Bound & $11.2$ & $81.20$ & $81.5$ & - & - \\
			\midrule
			UCIR\cite{hou2019learning} & $11.2$ & $68.09$ & $57.3$ & -& - \\
			PODNet\cite{douillard2020podnet} & $11.2$ & $74.33$ & - & -& -\\
			TPCIL\cite{tao2020tpcil} & $11.2$ & $74.81$ & $66.91$ & - & - \\ 
			\midrule
			Ours(w/o P) & $67.20$ & $\textbf{78.20}$ & $\textbf{74.92}$   & $\textbf{94.20}$    & $\textbf{91.30}$ \\
			Ours & $\textbf{8.87}$  & $\textbf{77.73}$ & $\textbf{72.06}$ & $\textbf{94.01}$    & $\textbf{91.64}$ \\
	\bottomrule[0.3mm]
	\end{tabular}}
%.  
	\end{minipage}}
	\caption{\fontsize{10}{12}\selectfont{Results on ImageNet-100  and ImageNet-1000 datasets.\textbf{Left}: The results on ImageNet100-B0 and ImageNet1000-B0 benchmark. \textbf{Right}: The results on ImageNet100-B50 benchmark. \emph{$\#$Paras} means the average number of parameters during inference over steps, which is counted by million. \emph{Avg} means the average accuracy ($\%$) over steps. \emph{Last} is the accuracy ($\%$) of the last step. \emph{Ours(w/o P)} means our method without pruning.}}
	\label{tab:imagenet}
\end{table*}
%\begin{figure*}[t]
%	\centering
%	% \begin{minipage}{0.25\textwidth}
%	% \begin{minipage}{0.25\textwidth}
%	\hspace{-4mm}
%	\begin{subfigure}{0.24\textwidth}
%		\includegraphics[width=\textwidth]{figures/step_perf/CIFAR100_BASE0_20S.pdf}
%		\caption{CIFAR100-B0 20 steps}
%		% \end{minipage}
%	\end{subfigure}
%	\hspace{-2mm}
%	% \begin{minipage}{0.25\textwidth}
%	\begin{subfigure}{0.24\textwidth}
%		\includegraphics[width=\textwidth]{figures/step_perf/CIFAR100_BASE0_50S.pdf}
%		\caption{CIFAR100-B0 50 steps}
%		% \end{minipage}
%	\end{subfigure}
%	\hspace{-2mm}
%	% \begin{minipage}{0.33\textwidth}
%	\begin{subfigure}{0.24\textwidth}
%		\includegraphics[width=\textwidth]{figures/step_perf/CIFAR100_BASE50_5S.pdf}
%		\caption{CIFAR100-B50 5 steps}
%		% \end{minipage}
%	\end{subfigure}
%	\hspace{-2mm}
%	% \begin{minipage}{0.33\textwidth}
%	\begin{subfigure}{0.24\textwidth}
%		\includegraphics[width=\textwidth]{figures/step_perf/CIFAR100_BASE50_10S.pdf}
%		\caption{CIFAR100-B50 10 steps}
%		% \end{minipage}
%	\end{subfigure}
%	\caption{\fontsize{10}{12}\selectfont{The performance for each step. \textbf{Left} is evaluated on CIFAR100-B0 of 20 and 50 steps and \textbf{Right} is evaluated on CIFAR100-B50 of 5 and 10 steps.}}\vspace{-2mm}
%	\label{fig:CIFAR100_steps}
%\end{figure*}

\begin{figure*}[t]
	\centering
%	\vspace{-3.5mm}
	\includegraphics[width=\textwidth]{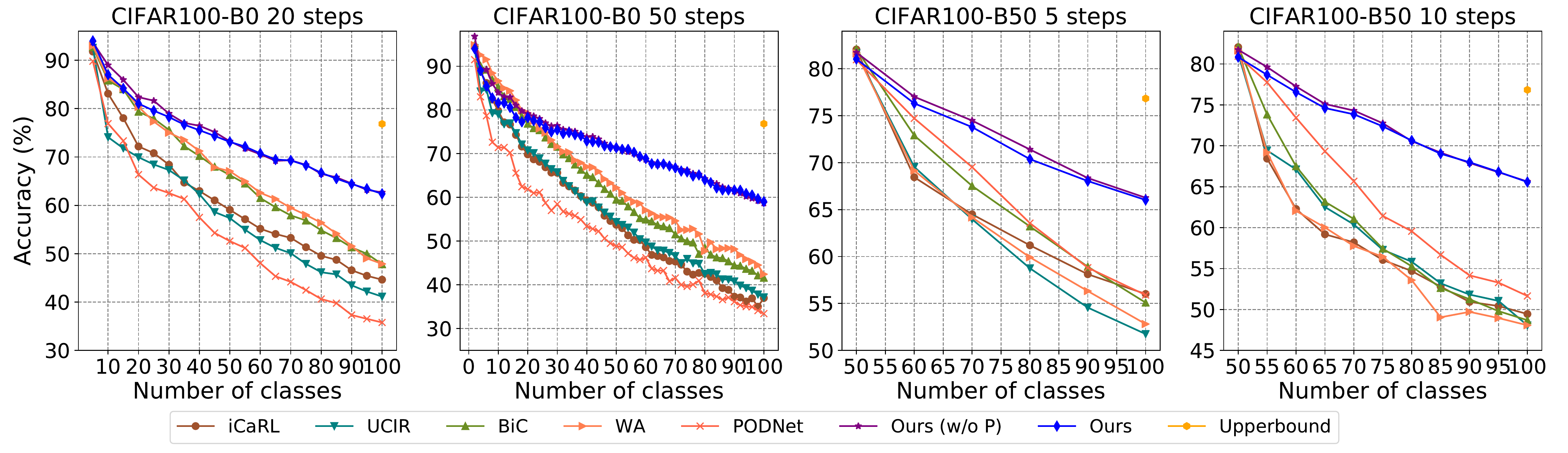}
	\vspace{-6.5mm}
	\caption{\fontsize{10}{12}\selectfont{The performance for each step. \textbf{Left} is evaluated on CIFAR100-B0 of 20 and 50 steps and \textbf{Right} is evaluated on CIFAR100-B50 of 5 and 10 steps.}}\vspace{-2mm}
	\label{fig:CIFAR100_steps}
\end{figure*}

\paragraph{Datasets}
CIFAR-100~\cite{krizhevsky2009learning} consists of 32x32 pixel color images with 100 classes. It contains 50,000 images for training with 500 images per class, and 10,000 images for evaluation with 100 images per class.
ImageNet-1000~\cite{deng2009imagenet} is a large-scale dataset from 1,000 classes which includes about 1.2 million RGB images for training and 50,000 images for validation.
ImageNet-100~\cite{rebuffi2017icarl,hou2019learning} is built by selecting 100 classes from the ImageNet-1000 dataset.

\vspace{-1mm}
\paragraph{Benchmark Protocols}
% Protocol 1) simulates the environment changes fast/much
% Protocol 2) simulates incremental learning starting with the pretrained model
For the CIFAR-100 benchmark, we test our methods on two popular protocols including \textit{1)CIFAR100-B0:} we follow the protocol proposed in~\cite{rebuffi2017icarl}, which trains all 100 classes in several splits including 5, 10, 20, 50 incremental steps with fixed memory size of 2,000 exemplars over batches;
% \modify{incremental steps of 5,10,20,50 classes at a time}
\textit{2)CIFAR100-B50:} we follow the protocol introduced in~\cite{hou2019learning}, which starts from a model trained on 50 classes, and the remaining 50 classes are divided into splits of 2, 5, and 10 steps with 20 examples as memory per class.
We compare the top-1 average incremental accuracy which takes the average of the accuracy for each step.

We also evaluate our method on ImageNet-100 with two protocols that are \textit{1)ImageNet100-B0:} the protocol~\cite{rebuffi2017icarl} trains the model in batches of 10 classes from scratch with fixed memory size 2,000 over batches; \textit{2)ImageNet100-B50:} the protocol~\cite{hou2019learning} starts from a model trained on 50 classes, and the remaining 50 classes come in 10 steps with 20 examples per class as memory.
For the sake of fairness, we use the same ImageNet subset and class order following the protocols~\cite{rebuffi2017icarl,hou2019learning}.
For ImageNet-1000, we evaluate our method on the protocol~\cite{rebuffi2017icarl}, known as \textit{ImageNet1000-B0} benchmark, that trains the model in batches of 100 classes with 10 steps in total and set the fixed memory size as 20,000.
Detailedly, we use the same class order as~\cite{rebuffi2017icarl} for ImageNet-1000.
Moreover, we compare the top-1 and top-5 average incremental accuracy and the last step accuracy on ImageNet-100 and ImageNet-1000 datasets.

\vspace{-3mm}

\paragraph{Implementation Details}
Our method is implemented with PyTorch~\cite{paszke2017automatic}.
%and the implementation code will be made publicly available.
For CIFAR-100, we adopt ResNet-18 as feature extractor $\mathcal{F}_t$ following RPSNet~ \cite{rajasegaran2019random}.
We note that most previous works use a modified 32-layers ResNet\cite{rebuffi2017icarl}, which has fewer channels and residual blocks compared to standard ResNet-32.
We argue that such a small network is not suitable because it cannot achieve competitive results on CIFAR100 compared with standard 18-layers ResNet\cite{he2016deep} and may underestimate the performance of methods.
We run these methods with standard ResNet-18 on the same class orders based on their code implementation. For those without releasing the codes, we report the results based on our implementation. For RPSNet, we use the results in their paper directly.
For ImageNet-100 and ImageNet-1000 benchmarks, we use 18-layers ResNet as the basic network.
In these experiemnts, we select exemplars as memory based on the herding selection strategy\cite{Welling09} following the previous works\cite{rebuffi2017icarl}.
Furthermore, we run experiments on three different class orders and report average$\pm$standard deviations in the results.
We also provide the experimental results on CIFAR-100 based on modified 32-layers ResNet \cite{rebuffi2017icarl} in the appendix, which proves the superiority of our method again.
%could serve as sanity-check of our  for other methods and
We follow the protocol in \cite{douillard2020podnet, serra2018overcoming} and tune the hyper-parameters on a validation set created by holding out a part of original training data.
The details of the hyperparameters are added to the appendix.

\subsection{Evaluation on CIFAR100}\label{subsec:cifar100}

\paragraph{Quantitative Results}
Table~\ref{tab:cifar100_base0} summarizes the results of CIFAR100-B0 benchmark.
We can see that our method consistently outperforms other methods by a sizable margin at different incremental splits.
As the number of steps increases in the split, it is observed that the margin between our method and other methods continuously increases which indicates that our method performs better on the difficult splits with longer steps.
Particularly, under the incremental setting of 50 steps, we improve the average incremental accuracy from $64.32\%$ to $72.05\%$($+\textbf{7.73\%}$) of ours with fewer parameters.
It is worth noting that although huge model parameters are reduced, the performance degradation of our method caused by pruning can be ignored, which demonstrates that the success of our pruning method.
As shown in the left panel of Figure~\ref{fig:CIFAR100_steps}, it is observed that our method consistently surpasses other methods at every step for different splits.
Moreover, the gap between our method and other methods increases with the continuous adding of novel classes.
Specifically, under the incremental split of 50 steps, the last step accuracy is boosted from $42.75\%$ to $58.66\%$($+\textbf{15.91\%}$), which further proves that the effectiveness of our method.
% In particular, our method improves from 64.32 to 72.05(+7.73$\%$) compared to the SOTA 64.32 with less parameters.

We also compare the performance of our method with previous methods in Table~\ref{tab:cifar100_base50} on the CIFAR100-B50 benchmark, which shows our method improves the performance with a significant gain in all splits.
% \modify{We can see that our method outperforms other methods by a sizable margin in all splits.}
In particular, under the incremental setting of 10 steps, our method outperforms PODNet by $\textbf{8.41\%}$ average incremental accuracy.
As the right panel of Figure~\ref{fig:CIFAR100_steps} shows, our method performs better than other methods at each step for all splits.
Especially, our method improves from $52.56\%$ to $65.58\%$($+\textbf{13.02\%}$) for the last step accuracy in the split of 10 steps.
Moreover, our method achieves similar performance with much fewer parameters compared to our method without pruning.
\begin{figure*}[t]
	\centering
	\includegraphics[width=.90\textwidth]{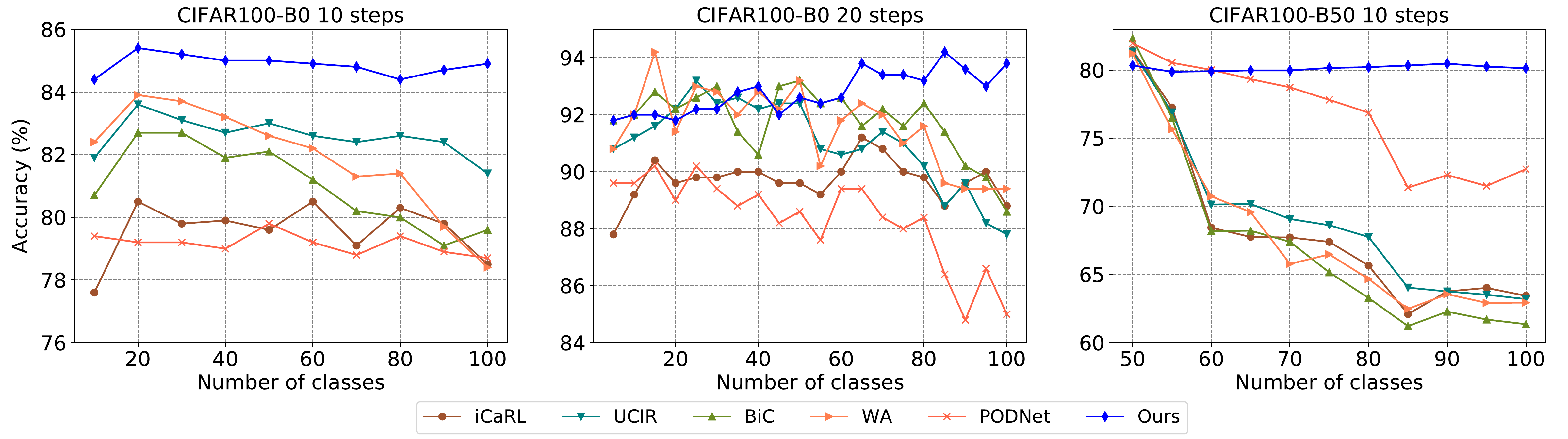}
	\vspace{-2.5mm}
	\caption{\fontsize{10}{12}\selectfont{\textbf{Analysis}. The backward transfer of representation by observing the changes of $A_{\mathcal{Y}_1}^t$ for different splits.}}
	\label{fig:forgetting}
\end{figure*}

It is worth noting that previous methods often perform well only on one of the protocols where WA is the state-of-the-art on CIFAR100-B0 and PODNet is state-of-the-art on CIFAR100-B50.
By contrast, our method consistently surpasses other methods on both protocols.
%\textcolor{red}{Besides, as Table \ref{tab:cifar100_base0} \& \ref{tab:cifar100_base50} show, our method is not order-sensitive as the standard deviation of the average accuracy is small.}
%the standard deviation of the average accuracy is consistently lower than 0.9\%
\vspace{-3mm}
\paragraph{The effects of model size}%also
We conduct extensive experiments to study the effect of model size on performance.
As shown in Figure~\ref{fig:cifar_model_size}, we can see that our method consistently and significantly performs better than other methods at various model sizes.% and incremental learning steps.
We also note that the improvement of our method compared to most other methods becomes more significant with the increasing of model size, which illustrates that our method can exploit the potential of a large model.

\subsection{Evaluation on ImageNet}\label{subsec:imagenet}
Table~\ref{tab:imagenet} summarizes the experimental results for the ImageNet-100 and ImageNet-1000 datasets.
We can see that our method consistently surpasses other methods with a considerable margin for all splits on ImageNet-100 and ImageNet-1000 datasets, especially the last step accuracy.
Specifically, our method outperforms the state-of-the-art with about $\textbf{1.79\%}$ for the average top-5 accuracy on the ImageNet100-B0 benchmark.
For ImageNet100-B50 benchmark, the last step top-1 accuracy is improved from $66.91\%$ to $72.06\%$(+$\textbf{5.15\%}$).
Furthermore, our method improves the final step top-1 accuracy from $55.6\%$ to $58.62\%$($+\textbf{3.02\%})$ on ImageNet1000-B0 benchmark. 
While the top-5 accuracy gap is smaller, we believe it is because top-5 accuracy is more tolerant to slightly inaccurate predictions and thus less sensitive to forgetting.

\begin{table}[t]
	\vspace{-2mm}
	\footnotesize
	\centering
	\vspace{-1mm}
	\fontsize{11}{12}\selectfont{
		\resizebox{0.27\textwidth}{!}{
			\begin{tabular}{cccc}
				\toprule[0.3mm]
				\multicolumn{2}{c}{\textbf{Components}} & \multirow{2}{*}{\textbf{Avg}} & \multirow{2}{*}{\textbf{Last}}           \\
				\textbf{E.R.}            & \textbf{Aux.}                 &                        &         \\
				\hline
				\xmark             & \xmark               & $61.84$                & $40.81$ \\
				\checkmark         & \xmark               & $73.26$                & $63.07$ \\
				\checkmark         & \checkmark           & $75.36$                & $65.34$ \\
				\bottomrule[0.3mm]
			\end{tabular}}}
	\caption{\fontsize{10}{12}\selectfont{The contribution of each component. \emph{E.R.} means expandable representation. \emph{Aux.} means using auxiliary loss.}}
	\label{tab:ablation}
	\vspace{-4mm}
\end{table}

\subsection{Ablation Study and Analysis}\label{subsec:ablation}
%In this subsection, 
We conduct exhaustive ablation study to evaluate the contribution of each component for our method. 
We also conduct sensitive study for hyper-parameters stated in the appendix. %and results are provided
Moreover, we study the backward transfer and forward transfer of the representation for each method.
\vspace{-3.5mm}
\paragraph{The effect of each component}
Table~\ref{tab:ablation} summarizes the results of our ablative experiments on CIFAR100-B0 with 10 steps.
We can see that the average accuracy is improved significantly from $61.84\%$ to $73.26\%$ by representation expansion.
We also show that the performance of the model is further improved with $\textbf{2.10\%}$ gain using auxiliary loss.

\vspace{-3mm}
\paragraph{Backward Transfer for Representation}
% Measure

To assess the quality of representation, we introduce an ideal decision boundary obtained by finetuning the classifier with all observed data, which allows us to exclude the influence of the classifier.
%within the label space $\mathcal{Y}_k $
We then define classification accuracy $A^t_{\mathcal{Y}_k}$ at step t as the accuracy on the test images of class set $\mathcal{Y}_k$, where the prediction space of the model is restricted to $\mathcal{Y}_k$. By observing the $A^t_{\mathcal{Y}_k}$ curve over $t$, we can see how the representation quality evolves along the increments. Figure~\ref{fig:forgetting} shows the results of  CIFAR100-B0 with 10 incremental steps.    
% by suppressing the probability outputs not in $\mathcal{Y}_k$
%We measure the backward transfer of representation on by observing the $A^t_{\mathcal{Y}_k}$ curve, which is shown in .
%To assess the backward transfer for representation for different methods, the accuracy $A^t_{\mathcal{Y}_k}$ at step $t$ is evaluated on the test set containing only images of classes $\mathcal{Y}_k$ where the classifier is finetuned with all observed data and the classification space is restricted to $\mathcal{Y}_t$ by suppressing the outputs of the classes not in $\mathcal{Y}_k$.
% Observation %, shown in Table ~\ref{tab:FWT}, 
We also compute a backward transfer value for different methods as follows:
\vspace{-3mm}
\begin{equation}
	\text{BWT} = \frac{1}{T-1}\sum^{T}_{i=2}\frac{1}{i}\sum_{j=1}^{i} A^{i}_{\mathcal{Y}_{j}} - A^{j}_{\mathcal{Y}_{j}}
\vspace{-2.5mm}
\end{equation}
The results are shown in Table \ref{tab:FWT}.
We can see that other methods suffer from severe forgetting. %catastrophic
In contrast, our method even achieves positive backward transfer $\textbf{+1.36\%}$ and the accuracy increases with respect to steps, which further proves the superiority of our method.
\begin{table}[t]
	\vspace{-2mm}
	\footnotesize
	\centering
	\fontsize{12}{12}\selectfont{
		\resizebox{0.47\textwidth}{!}{
			\begin{tabular}{ccccccc}
				\toprule[0.3mm]
				\textbf{Methods}    & \textbf{iCaRL}   & \textbf{UCIR}    & \textbf{BiC}     & \textbf{WA}      & \textbf{PODNet}   & \textbf{Ours}             \\
				\hline
				BWT ($\%$) & $-4.14$ & $-8.52$ & $-3.40$ & $-3.18$ & $-16.27$ & $\textbf{+1.36}$ \\
				FWT ($\%$) & $-4.91$ & $-5.56$ & $-0.17$ & $+0.82$ & $-5.58$  & $\textbf{+1.49}$ \\
				\bottomrule[0.3mm]
			\end{tabular}}}
	\caption{\fontsize{10}{12}\selectfont{Backward transfer and Forward transfer (FWT) for representation.}}\vspace{-3mm}
	\label{tab:FWT}
	\vspace{-1.5mm}
\end{table}
%the union $\mathcal{Y}=\mathcal{Y}_i \cup \mathcal{Y}_j$ of label space of step $i,j$ by suppressing the outputs out of the label space $\mathcal{Y}$.

\vspace{-3.5mm}
\paragraph{Forward Transfer for Representation}
We also measure the influence of existing knowledge on the performance of subsequent concepts on CIFAR100-B0 with 10 incremental steps, known as forward transfer.
Specifically, we define a forward transfer rate for representation as follows
\vspace{-3mm}
\begin{equation}
	\text{FWT} = \frac{1}{T-1}\sum_{i=2}^{T} A^{i}_{\mathcal{Y}_{i}} - \bar{A}^{i}_{\mathcal{Y}_{i}}
	\vspace{-2.5mm}
\end{equation}
where $\bar{A}^{i}_{\mathcal{Y}_{i}}$ is the test accuracy obtained by model trained on available data $\tilde{\mathcal{D}_{t}}$ with only cross-entropy loss at random initialization.
As shown in Table~\ref{tab:FWT}, it is observed that most methods have negative forward transfer, which indicates that they sacrifice the flexibility of adaptation to novel concepts.
By contrast, our method achieves $\textbf{+1.49\%}$ FWT which implies that our method not only makes the model highly flexible but also brings the positive forward transfer.

\section{Conclusion}\label{sec:conclusion}
\vspace{-1.5mm}
In this work, we propose dynamically expandable representation to improve the representation for class incremental learning.
%a simple and yet effective solution focus on 
At each step, we freeze previously learned representation and augment it with novel parameterized feature. 
We also introduce channel-level mask-based pruning to dynamically expand representation according to the difficulty of novel concepts and an auxiliary loss to learn the novel discriminative features better. %method  a 
%Moreover, we introduce . %
%To reduce the catastrophic forgetting and achieve transfer, we build the representation by concatenating the previously learned discriminative features for old concepts and new features where the previous features are invariant by freezing the corresponding parameters.
%To remove the model redundancy and keep model compact, we apply a channel-level pruning method using differentiable mask for novel features.
We conduct exhaustive experiments on the three major incremental classification benchmarks. %including CIFAR-100, ImageNet-100, and ImageNet-1000.
The experimental results show that our method consistently performs better than other methods with a sizable margin. % demonstrate  
Interestingly, we also find that our method can even achieve positive backward and forward transfer.% 

{\small
    \bibliographystyle{ieee_fullname}
    \bibliography{egbib}

\begin{thebibliography}{10}\itemsep=-1pt

\bibitem{abatiTBCCB2020ccgn}
Davide Abati, Jakub Tomczak, Tijmen Blankevoort, Simone Calderara, Rita
  Cucchiara, and Babak~Ehteshami Bejnordi.
\newblock Conditional channel gated networks for task-aware continual learning.
\newblock In {\em Proceedings of the IEEE conference on computer vision and
  pattern recognition (CVPR)}, 2020.

\bibitem{aljundi2018memory}
Rahaf Aljundi, Francesca Babiloni, Mohamed Elhoseiny, Marcus Rohrbach, and
  Tinne Tuytelaars.
\newblock Memory aware synapses: Learning what (not) to forget.
\newblock In {\em Proceedings of the European Conference on Computer Vision
  (ECCV)}, 2018.

\bibitem{castro2018eeil}
Francisco~M. Castro, Manuel~J. Mar{\'{\i}}n{-}Jim{\'{e}}nez, Nicol{\'{a}}s
  Guil, Cordelia Schmid, and Karteek Alahari.
\newblock End-to-end incremental learning.
\newblock In {\em Proceedings of the European Conference on Computer Vision
  (ECCV)}, 2018.

\bibitem{chaudhry2018riemannian}
Arslan Chaudhry, Puneet~K Dokania, Thalaiyasingam Ajanthan, and Philip~HS Torr.
\newblock Riemannian walk for incremental learning: Understanding forgetting
  and intransigence.
\newblock In {\em Proceedings of the European Conference on Computer Vision
  (ECCV)}, 2018.

\bibitem{deng2009imagenet}
Jia Deng, Wei Dong, Richard Socher, Li-Jia Li, Kai Li, and Li Fei-Fei.
\newblock Imagenet: A large-scale hierarchical image database.
\newblock In {\em Proceedings of the IEEE conference on computer vision and
  pattern recognition (CVPR)}, 2009.

\bibitem{douillard2020podnet}
Arthur Douillard, Matthieu Cord, Charles Ollion, Thomas Robert, and Eduardo
  Valle.
\newblock Podnet: Pooled outputs distillation for small-tasks incremental
  learning.
\newblock In {\em Proceedings of the European Conference on Computer Vision
  (ECCV)}, 2020.

\bibitem{fernando2017pathnet}
Chrisantha Fernando, Dylan Banarse, Charles Blundell, Yori Zwols, David Ha,
  Andrei~A Rusu, Alexander Pritzel, and Daan Wierstra.
\newblock Pathnet: Evolution channels gradient descent in super neural
  networks.
\newblock {\em arXiv preprint arXiv:1701.08734}, 2017.

\bibitem{FrenchC02}
Robert~M. French and Nick Chater.
\newblock Using noise to compute error surfaces in connectionist networks: {A}
  novel means of reducing catastrophic forgetting.
\newblock {\em Neural Comput.}, 2002.

\bibitem{grossberg2013adaptive}
Stephen Grossberg.
\newblock Adaptive resonance theory: How a brain learns to consciously attend,
  learn, and recognize a changing world.
\newblock {\em Neural Networks}, 2013.

\bibitem{he2016deep}
Kaiming He, Xiangyu Zhang, Shaoqing Ren, and Jian Sun.
\newblock Deep residual learning for image recognition.
\newblock In {\em Proceedings of the IEEE conference on computer vision and
  pattern recognition (CVPR)}, 2016.

\bibitem{hinton15distill}
Geoffrey Hinton, Oriol Vinyals, and Jeffrey Dean.
\newblock Distilling the knowledge in a neural network.
\newblock In {\em Advances in Neural Information Processing Systems (NeurIPS)
  Workshop}, 2015.

\bibitem{hou2019learning}
Saihui Hou, Xinyu Pan, Chen~Change Loy, Zilei Wang, and Dahua Lin.
\newblock Learning a unified classifier incrementally via rebalancing.
\newblock In {\em Proceedings of the IEEE conference on computer vision and
  pattern recognition (CVPR)}, 2019.

\bibitem{hung2019cpg}
Steven C.~Y. Hung, Cheng{-}Hao Tu, Cheng{-}En Wu, Chien{-}Hung Chen, Yi{-}Ming
  Chan, and Chu{-}Song Chen.
\newblock Compacting, picking and growing for unforgetting continual learning.
\newblock In {\em Advances in Neural Information Processing Systems (NeurIPS)},
  2019.

\bibitem{IoffeS15bn}
Sergey Ioffe and Christian Szegedy.
\newblock Batch normalization: Accelerating deep network training by reducing
  internal covariate shift.
\newblock In {\em International Conference on Machine Learning (ICML)}, 2015.

\bibitem{KangXRYGFK20decouple}
Bingyi Kang, Saining Xie, Marcus Rohrbach, Zhicheng Yan, Albert Gordo, Jiashi
  Feng, and Yannis Kalantidis.
\newblock Decoupling representation and classifier for long-tailed recognition.
\newblock In {\em International Conference on Learning Representations (ICLR)},
  2020.

\bibitem{kirkpatrick2017overcoming}
James Kirkpatrick, Razvan Pascanu, Neil Rabinowitz, Joel Veness, Guillaume
  Desjardins, Andrei~A Rusu, Kieran Milan, John Quan, Tiago Ramalho, Agnieszka
  Grabska-Barwinska, et~al.
\newblock Overcoming catastrophic forgetting in neural networks.
\newblock {\em Proceedings of the national academy of sciences (PNAS)}, 2017.

\bibitem{krizhevsky2009learning}
Alex Krizhevsky and Geoffrey Hinton.
\newblock Learning multiple layers of features from tiny images.
\newblock {\em Technical report, University of Toronto}, 2009.

\bibitem{lee2017overcoming}
Sang-Woo Lee, Jin-Hwa Kim, Jaehyun Jun, Jung-Woo Ha, and Byoung-Tak Zhang.
\newblock Overcoming catastrophic forgetting by incremental moment matching.
\newblock In {\em Advances in neural information processing systems (NeurIPS)},
  2017.

\bibitem{li2017incremental}
Lufan Li, Zhang Jun, Jiawei Fei, and Shuohao Li.
\newblock An incremental face recognition system based on deep learning.
\newblock In {\em International Conference on Machine Vision Applications
  (MVA)}, 2017.

\bibitem{li2019learn}
Xilai Li, Yingbo Zhou, Tianfu Wu, Richard Socher, and Caiming Xiong.
\newblock Learn to grow: A continual structure learning framework for
  overcoming catastrophic forgetting.
\newblock In {\em International Conference on Machine Learning(ICML)}, 2019.

\bibitem{mallya2018piggyback}
Arun Mallya, Dillon Davis, and Svetlana Lazebnik.
\newblock Piggyback: Adapting a single network to multiple tasks by learning to
  mask weights.
\newblock In {\em Proceedings of the European Conference on Computer Vision
  (ECCV)}, 2018.

\bibitem{mallya2018packnet}
Arun Mallya and Svetlana Lazebnik.
\newblock Packnet: Adding multiple tasks to a single network by iterative
  pruning.
\newblock In {\em Proceedings of the IEEE conference on computer vision and
  pattern recognition (CVPR)}, 2018.

\bibitem{ostapenko2019learning}
Oleksiy Ostapenko, Mihai Puscas, Tassilo Klein, Patrick Jahnichen, and Moin
  Nabi.
\newblock Learning to remember: A synaptic plasticity driven framework for
  continual learning.
\newblock In {\em Proceedings of the IEEE Conference on Computer Vision and
  Pattern Recognition (CVPR)}, 2019.

\bibitem{paszke2017automatic}
Adam Paszke, Sam Gross, Soumith Chintala, Gregory Chanan, Edward Yang, Zachary
  DeVito, Zeming Lin, Alban Desmaison, Luca Antiga, and Adam Lerer.
\newblock Automatic differentiation in pytorch.
\newblock 2017.

\bibitem{pierre2018incremental}
John~M Pierre.
\newblock Incremental lifelong deep learning for autonomous vehicles.
\newblock In {\em International Conference on Intelligent Transportation
  Systems (ITSC)}, 2018.

\bibitem{rajasegaran2019random}
Jathushan Rajasegaran, Munawar Hayat, Salman~H Khan, Fahad~Shahbaz Khan, and
  Ling Shao.
\newblock Random path selection for continual learning.
\newblock In {\em Advances in Neural Information Processing Systems (NeurIPS)},
  2019.

\bibitem{rebuffi2017icarl}
Sylvestre-Alvise Rebuffi, Alexander Kolesnikov, Georg Sperl, and Christoph~H
  Lampert.
\newblock icarl: Incremental classifier and representation learning.
\newblock In {\em Proceedings of the IEEE conference on computer vision and
  pattern recognition (CVPR)}, 2017.

\bibitem{Robins93a}
Anthony~V. Robins.
\newblock Catastrophic forgetting in neural networks: the role of rehearsal
  mechanisms.
\newblock In {\em International Two-Stream Conference on Artificial Neural
  Networks and Expert Systems, {ANNES}}, 1993.

\bibitem{Robins95}
Anthony~V. Robins.
\newblock Catastrophic forgetting, rehearsal and pseudorehearsal.
\newblock {\em Connect. Sci.}, 1995.

\bibitem{serra2018overcoming}
Joan Serra, Didac Suris, Marius Miron, and Alexandros Karatzoglou.
\newblock Overcoming catastrophic forgetting with hard attention to the task.
\newblock In {\em International Conference on Machine Learning (ICML)}, 2018.

\bibitem{thrun1995lifelong}
Sebastian Thrun and Tom~M Mitchell.
\newblock Lifelong robot learning.
\newblock {\em Robotics and autonomous systems}, 1995.

\bibitem{Welling09}
Max Welling.
\newblock Herding dynamical weights to learn.
\newblock In {\em International Conference on Machine Learning (ICML)}, 2009.

\bibitem{wu2019bic}
Yue Wu, Yinpeng Chen, Lijuan Wang, Yuancheng Ye, Zicheng Liu, Yandong Guo, and
  Yun Fu.
\newblock Large scale incremental learning.
\newblock In {\em Proceedings of the IEEE conference on computer vision and
  pattern recognition (CVPR)}, 2019.

\bibitem{tao2020tpcil}
Tao Xiaoyu, Chang Xinyuan, Hong Xiaopeng, Wei Xing, and Gong Yihong.
\newblock Topology-preserving class-incremental learning.
\newblock In {\em Proceedings of the European Conference on Computer Vision
  (ECCV)}, 2020.

\bibitem{yoon2018lifelong}
Jaehong Yoon, Eunho Yang, Jeongtae Lee, and Sung~Ju Hwang.
\newblock Lifelong learning with dynamically expandable networks.
\newblock In {\em International Conference on Learning Representations (ICLR)},
  2018.

\bibitem{yu2020semantic}
Lu Yu, Bartlomiej Twardowski, Xialei Liu, Luis Herranz, Kai Wang, Yongmei
  Cheng, Shangling Jui, and Joost van~de Weijer.
\newblock Semantic drift compensation for class-incremental learning.
\newblock In {\em Proceedings of the IEEE Conference on Computer Vision and
  Pattern Recognition (CVPR)}, 2020.

\bibitem{zenke2017continual}
Friedemann Zenke, Ben Poole, and Surya Ganguli.
\newblock Continual learning through synaptic intelligence.
\newblock In {\em International Conference on Machine Learning (ICML)}, 2017.

\bibitem{zhang2018heated}
Xu Zhang, Felix~Xinnan Yu, Svebor Karaman, Wei Zhang, and Shih-Fu Chang.
\newblock Heated-up softmax embedding.
\newblock {\em arXiv preprint arXiv:1809.04157}, 2018.

\bibitem{zhao2020wa}
Bowen Zhao, Xi Xiao, Guojun Gan, Bin Zhang, and Shu-Tao Xia.
\newblock Maintaining discrimination and fairness in class incremental
  learning.
\newblock In {\em Proceedings of the IEEE conference on computer vision and
  pattern recognition (CVPR)}, 2020.

\end{thebibliography}
}
\clearpage

\begin{appendices}
\section{Hyperparameters}
\paragraph{Representation learning stage} For CIFAR-100, we use SGD to train the model with batch size 128, weight decay 0.0005. We adopt the warmup strategy with the ending learning rate 0.1 for 10 epochs. After the warmup, we run the SGD with 160 epochs and the learning rate decays at 100, 120 epochs with 0.1.
For ImageNet100 and ImageNet1000, we adopt SGD with batch size 256, weight decay 0.0005. We also use the same warmup strategy as in CIFAR100. After the warmup, model is trained for 120 epochs. Learning rate starts from 0.1 and decays by 0.1 rates after 30, 60, 80, and 90 epochs.

For the coefficients in the loss function, $\lambda_{a}$ is set to 1 for all experiments in the paper. 
$\lambda_s$ is tuned to ensure our learned model has comparable number of parameters to other methods for fair comparison.
Moreover, for simplicity, we set the same $\lambda_s$ for every steps in each experiment. 
%The coefficient of sparsity loss $\lambda_s$ is set differently in different experiments so that our method use less parameters on average than other methods. 
For experiments of CIFAR100-B0, $\lambda_s$ is set as 0.75. For experiments of CIFAR100-B50, $\lambda_s$ is set as 0.25. For experiments on ImageNet100 and ImageNet, $\lambda_s$ is set as 0.75 for ImageNet100-B0, 0.5 for ImageNet100-B50 and 0.75 for ImageNet.

\paragraph{Classifier learning stage} We adopt SGD optimizer with weight decay 0.0005 to update the classifier only for 30 epochs with SGD optimizer. Learning rate is 0.1 and decays with 0.1 rate at 15 epochs.
The temperature of cross-entropy loss are set as $\delta=5$ for CIFAR-100 and $\delta=1$ for ImageNet-100 and ImageNet-1000.

\section{Sensitive Study of Hyper-parameters}
We conduct a sensitive study of our method on CVIFAR100-B0 10 steps with different $\lambda_a$. 
The results are shown in Table \ref{tab:sensitive_lambda_a}, which demonstrates our method is robust to $\lambda_a$.
We also conduct experiments for different $\lambda_s$, which is shown in the Figure 1 in the main body.

\begin{table}[h]
	\centering
	\fontsize{11}{13}\selectfont{
		\resizebox{0.48\textwidth}{!}{
			\begin{tabular}{l|ccccc}
				\toprule[0.3mm]
				$\lambda_a$ & 0.1 & 0.5 & 1 & 5 & 10 \\
				\hline
				Avg &  $74.12_{\pm{0.06}}$    &  $74.41_{\pm{0.16}}$   &  $74.64_{\pm{0.28}}$ & $74.52_{\pm{0.33}}$ & $73.54_{\pm{0.22}}$     \\
				\bottomrule[0.3mm]
	\end{tabular}}}
	\vspace{-3mm}
	\caption{Sensitive study on effects of $\lambda_a$}
	\label{tab:sensitive_lambda_a}
\end{table}

\section{The Quality of Decision Boundary}
In this section, our goal is to verify that the high-quality linear classifier can be obtained even re-learning the old classes' decision boundary with memory $\mathcal{M}$.
Specifically, we compare our method with an `ideal' strategy that uses all the previous data to train the classifier in the second stage. Such an upper bound achieves $76.14\pm0.80\%$ on the CIFAR100-B0 10 steps, which is only slightly higher than our method ($74.64\pm0.28\%$). We also observed similar results on the other benchmarks, which show the efficacy of our second-stage learning.  

\section{Latency}
Regarding inference latency, we conduct an experimental comparison on the ImageNet with GTX 1080Ti. 
Our method achieves 1.07ms/image, which is comparable to other baseline methods, such as BiC and WA, which are 0.99ms/image.

\section{Results for modified 32-layer ResNet}\label{sec:modres32}
Most works use a modified 32-layer ResNet following iCaRL\cite{rebuffi2017icarl}.
We also compare the results of our method with other methods based on the modified 32-layer ResNet.
The results on CIFAR100-B0 are shown in Table \ref{tab:cifar100_base0_modres32} and the results on CIFAR100-B50 are shown in Table~\ref{tab:cifar100_base50_modres32}.
The results of other methods are reported in their papers.
It can be found that our method still outperforms other methods on both CIFAR100-B0 and CIFAR100-B50 even with a small network like modified ResNet-32.

\section{More detailed results on CIFAR100}
\begin{figure*}[t]
	\centering
	\includegraphics[width=.95\textwidth]{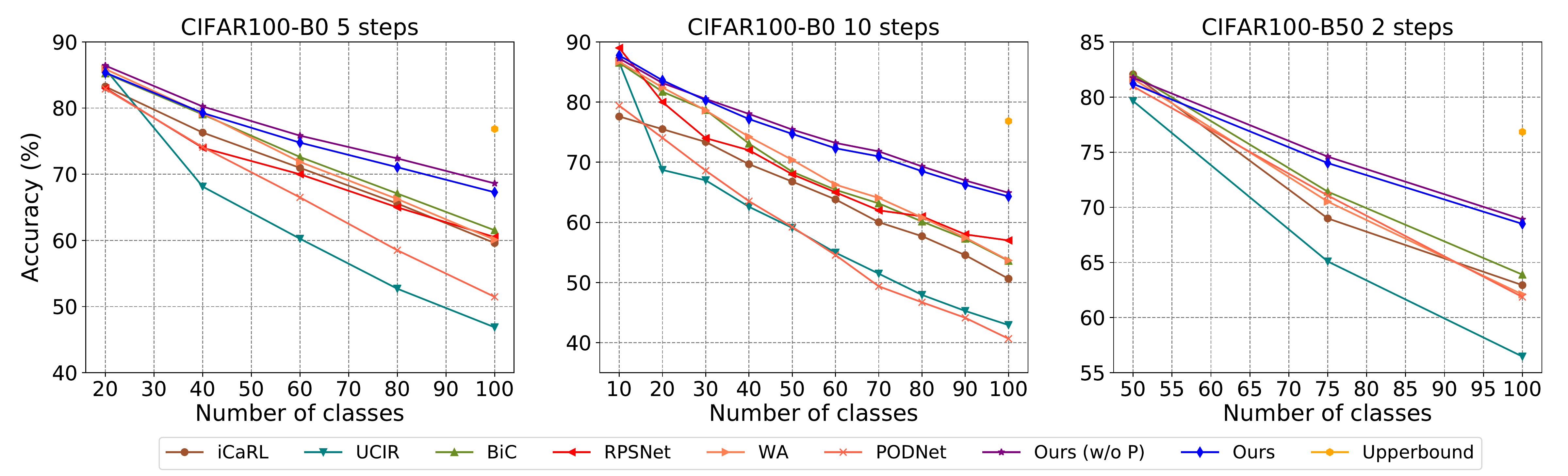}
	\vspace{-3.0mm}
	\caption{The performance for each step. \textbf{Left} is evaluated on CIFAR100-B0 of 5 steps. \textbf{Middle} is evaluated on CIFAR100-B0 of 10 steps. \textbf{Right} is evaluated on CIFAR100-B50 of 2 steps.}\vspace{-2mm}
	\label{fig:more_curve_cifar100}
\end{figure*}

Figure~\ref{fig:more_curve_cifar100} shows the performance with respect to steps on CIFAR100-B0 with 5 incremental steps and 10 incremental steps and CIFAR100-B0 with 2 incremental steps.
This also illustrates the superiority of our method.

\section{Detailed results on ImageNet}

\begin{figure*}[t]
	\centering
	\includegraphics[width=.95\textwidth]{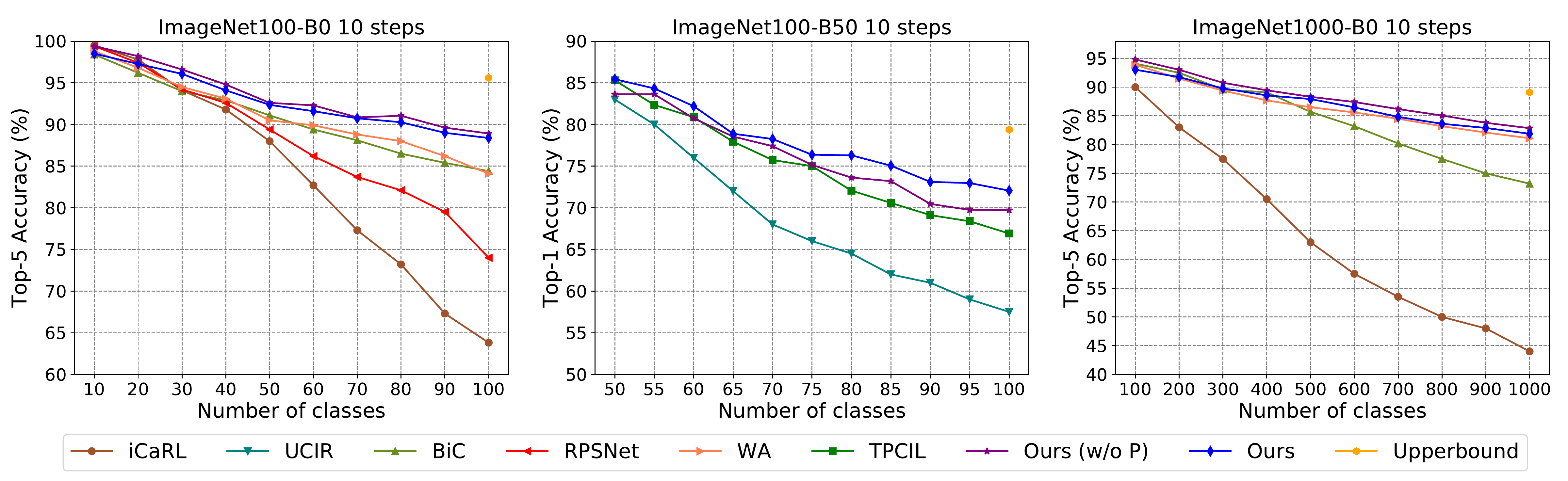}
	\vspace{-3.0mm}
	\caption{The performance for each step. \textbf{Left} is evaluated on ImageNet100-B0 of 10 steps. \textbf{Middle} is evaluated on ImageNet100-B50 of 10 steps. \textbf{Right} is evaluated on ImageNet1000-B0 of 10 steps.}
	\label{fig:more_curve_imagenet}
\end{figure*}

We also show the curves of performance with respect to steps on ImageNet100-B0, ImageNet100-B50 and ImageNet1000-B0 in Figure~\ref{fig:more_curve_imagenet}, which proves the effectiveness of our method on complex datasets.

\begin{table*}[t]
\centering
%\caption{Results on CIFAR100-B0 (average over 3 runs). \emph{$\#$Paras} means the average number of parameters used during inference over steps, whose unit is million parameters. \emph{Avg} means the average accuracy ($\%$) over steps. \emph{Ours} represents our method and \emph{Ours(w/o P)} means our method without pruning.}
\fontsize{10.53}{12}\selectfont{
\resizebox{0.9\textwidth}{!}{
\begin{tabular}{l|ll|ll|ll|ll}
%\centering
\toprule[0.3mm]
\multirow{2}{*}{\textbf{Methods}} &\multicolumn{2}{c|}{\textbf{$5$ steps}}       & \multicolumn{2}{c|}{\textbf{$10$ steps}}     & \multicolumn{2}{c|}{\textbf{$20$ steps}}    & \multicolumn{2}{c}{\textbf{$50$ steps}} \\
\cmidrule{2-9}
&                                \textbf{$\#$Paras}    & \textbf{Avg}  & \textbf{$\#$Paras}  & \textbf{Avg}     & \textbf{$\#$Paras}   & \textbf{Avg} & \textbf{$\#$Paras} & \textbf{Avg}    \\
\midrule
iCaRL~\cite{rebuffi2017icarl}       & $0.46$ & $67.20$ & $0.46$ & $64.04$ & $0.46$ & $61.16$ & $0.46$ & $57.00$ \\
BiC~\cite{hou2019learning}          & $0.46$ & $68.92$ & $0.46$ & $66.15$ & $0.46$ & $63.80$ & $0.46$ & -      \\
WA~\cite{zhao2020wa}                & $0.46$ & $70.00$ & $0.46$ & $67.25$ & $0.46$ & $64.33$ & $0.46$ & -      \\
\midrule
Ours(w/o P)                        & $1.38$ & $\textbf{73.00}\gain{3.00}$ & $2.53$ & $\textbf{71.29}\gain{4.04}$ & $4.83$ & $\textbf{71.07}\gain{6.74}$ & $11.73$ & $\textbf{70.58}\gain{13.58}$  \\
Ours                               & $0.42$ & $\textbf{72.31}\gain{2.31}$ & $0.52$ & $\textbf{69.41}\gain{2.16}$ & $0.45$ & $\textbf{68.82}\gain{4.49}$ & $0.70$ & $\textbf{67.29}\gain{10.29}$ \\
\bottomrule[0.3mm]
\end{tabular}}}
\caption{\fontsize{10}{12}\selectfont{Results on CIFAR100-B0 benchmark using modified 32-layer ResNet. \emph{$\#$Paras} means the average number of parameters used during inference over steps, which is counted by million. \emph{Avg} means the average accuracy ($\%$) over steps. \emph{Ours(w/o P)} means our method without pruning.}}
\label{tab:cifar100_base0_modres32}
%$\textbf{3}$$$$\color{gain}{(+3.7)}$
\end{table*}

\begin{table*}[h]
	\centering
	\fontsize{11.76}{12}\selectfont{
		\resizebox{0.75\textwidth}{!}{
			\footnotesize
			\begin{tabular}{l|ll|ll|ll}
				\toprule[0.3mm]
				\multirow{2}{*}{\textbf{Methods}}         & \multicolumn{2}{c|}{\textbf{2Steps}}   & \multicolumn{2}{c|}{\textbf{5Steps}}               & \multicolumn{2}{c}{\textbf{10Steps}}                                                                                                                          \\
				\cmidrule{2-7}
				                                 & \textbf{$\#$Paras}                     & \textbf{Avg}                                       & \textbf{$\#$Paras}                     & \textbf{Avg}                                       & \textbf{$\#$Paras}                     & \textbf{Avg}                                       \\
				\midrule
				UCIR~\cite{hou2019learning}       & $0.46$ & $66.76$ & $0.46$ & $63.42$ & $0.46$ & $60.18$ \\
				PODNet~\cite{douillard2020podnet} & -      & -       & $0.46$ & $64.83$ & $0.46$ & $64.03$ \\
				TPCIL~\cite{tao2020tpcil}         & - & -       & $0.46$ & $65.34$ & $0.46$ & $63.58$ \\
				\midrule
				Ours(w/o P)                      & $0.92$ & $\textbf{70.18}\gain{3.42}$ & $1.61$ & $\textbf{68.52}\gain{3.18}$ & $2.76$ & $\textbf{67.09}\gain{3.06}$ \\
				Ours                             & $0.32$ & $\textbf{69.52}\gain{2.76}$ & $0.59$ & $\textbf{67.60}\gain{2.26}$ & $0.61$ & $\textbf{66.36}\gain{2.33}$ \\
				\bottomrule[0.3mm]
			\end{tabular}}}
	\caption{\fontsize{10}{12}\selectfont{Results on CIFAR100-B50 using modified 32-layer ResNet. \emph{$\#$Paras} means the average number of parameters used during inference over steps, which is counted by million. \emph{Avg} means the average accuracy ($\%$) over steps. \emph{Ours(w/o P)} means our method without pruning.}} \vspace{-3mm}
	\label{tab:cifar100_base50_modres32}
\end{table*}

\end{appendices}

%\includepdf[pages=-]{incremental_classification.pdf}

\end{document}